\documentclass[review]{elsarticle}
\usepackage{amsmath, amssymb, eqparbox}
\usepackage{graphicx,psfrag,epsf}
\usepackage{enumerate}
\usepackage{url} 
\usepackage[margin=20mm]{geometry}
\usepackage{algorithm}
\usepackage{algpseudocode}
\usepackage{amsthm}
\usepackage{mathrsfs}
\usepackage{multirow}
\usepackage{pdfpages}
\usepackage{hyperref}
\usepackage{float}
\usepackage{xltabular}

\journal{Expert Systems with Applications}

\biboptions{authoryear}

\usepackage[hang,small,bf]{caption}
\usepackage[subrefformat=parens]{subcaption}
\theoremstyle{plain}
\newtheorem{thm}{Theorem}
\newtheorem*{thm*}{Theorem}
\newtheorem{cor}{Corollary}
\newtheorem*{cor*}{Corollary}

\theoremstyle{definition}

\floatstyle{plaintop}
\restylefloat{table}

\begin{document}
\begin{frontmatter}

\begin{titlepage}
\begin{center}
\vspace*{1cm}

\textbf{ \large A new type of federated clustering: A non-model-sharing approach}

\vspace{1.5cm}

Yuji Kawamata$^{a,c}$ (yjkawamata@gmail.com), Kaoru Kamijo$^b$ (s2520514@u.tsukuba.ac.jp), Masateru Kihira$^b$ (kihira.masateru.tg@alumni.tsukuba.ac.jp), Akihiro Toyoda$^b$ (toyoda.akihiro.as@alumni.tsukuba.ac.jp), Tomoru Nakayama$^b$ (nakayama.tomoru.tkb\_en@u.tsukuba.ac.jp), Akira Imakura$^{a,c}$ (imakura@cs.tsukuba.ac.jp), Tetsuya Sakurai$^{a,c}$ (sakurai@cs.tsukuba.ac.jp), Yukihiko Okada$^{a,c}$ (okayu@sk.tsukuba.ac.jp) \\

\hspace{10pt}

\begin{flushleft}
\small  
$^a$ Center for Artificial Intelligence Research, Tsukuba Institute for Advanced Research, University of Tsukuba, 1-1-1 Tennodai, Tsukuba, Ibaraki 305-8577, Japan \\
$^b$ Graduate School of Science and Technology, University of Tsukuba, 1-1-1 Tennodai, Tsukuba, Ibaraki 305-8573, Japan \\
$^c$ Institute of Systems and Information Engineering, University of Tsukuba, 1-1-1 Tennodai, Tsukuba, Ibaraki 305-8573, Japan

\vspace{1cm}
\textbf{Corresponding Author:} \\
Yuji Kawamata \\
Center for Artificial Intelligence Research, Tsukuba Institute for Advanced Research / Institute of Systems and Information Engineering, University of Tsukuba, 1-1-1 Tennodai, Tsukuba, Ibaraki 305-8577, Japan \\
Email: yjkawamata@gmail.com

\end{flushleft}        
\end{center}
\end{titlepage}

\title{A new type of federated clustering: A non-model-sharing approach}

\author[inst_cair,inst_feis]{Yuji Kawamata\corref{cor1}\fnref{equal}}
\ead{yjkawamata@gmail.com}

\author[inst_gs]{Kaoru Kamijo\fnref{equal}}
\ead{s2520514@u.tsukuba.ac.jp}

\author[inst_gs]{Masateru Kihira}
\ead{kihira.masateru.tg@alumni.tsukuba.ac.jp}

\author[inst_gs]{Akihiro Toyoda}
\ead{toyoda.akihiro.as@alumni.tsukuba.ac.jp}

\author[inst_gs]{Tomoru Nakayama}
\ead{nakayama.tomoru.tkb_en@u.tsukuba.ac.jp}

\author[inst_cair,inst_feis]{Akira Imakura}
\ead{imakura@cs.tsukuba.ac.jp}

\author[inst_cair,inst_feis]{Tetsuya Sakurai}
\ead{sakurai@cs.tsukuba.ac.jp}

\author[inst_cair,inst_feis]{Yukihiko Okada}
\ead{okayu@sk.tsukuba.ac.jp}

\cortext[cor1]{Corresponding author.}
\fntext[equal]{These authors contributed equally to this work.}
\address[inst_cair]{Center for Artificial Intelligence Research, Tsukuba Institute for Advanced Research, University of Tsukuba, 1-1-1 Tennodai, Tsukuba, Ibaraki 305-8577, Japan}
\address[inst_gs]{Graduate School of Science and Technology, University of Tsukuba, 1-1-1 Tennodai, Tsukuba, Ibaraki 305-8573, Japan}
\address[inst_feis]{Institute of Systems and Information Engineering, University of Tsukuba, 1-1-1 Tennodai, Tsukuba, Ibaraki 305-8573, Japan}

\begin{abstract}
In recent years, the growing need to leverage sensitive data across institutions has led to increased attention on Federated Learning (FL), a decentralized machine learning paradigm that enables model training without sharing raw data. However, existing FL-based clustering methods, known as Federated Clustering, typically assume simple data partitioning scenarios such as horizontal or vertical splits, and cannot handle more complex distributed structures. This study proposes Data Collaboration Clustering (DC-Clustering), a novel federated clustering method that supports clustering over complex data partitioning scenarios where horizontal and vertical splits coexist. In DC-Clustering, each institution shares only intermediate representations instead of raw data, ensuring privacy preservation while enabling collaborative clustering. The method allows flexible selection between k-means and spectral clustering, and achieves final results with a single round of communication with the central server. We conducted extensive experiments using synthetic and open benchmark datasets. The results show that our method achieves clustering performance comparable to central clustering where all data are pooled. DC-Clustering addresses an important gap in current FL research by enabling effective knowledge discovery from distributed heterogeneous data. Its practical properties—privacy preservation, communication efficiency, and flexibility—make it a promising tool for privacy-sensitive domains such as healthcare and finance.
\end{abstract}


\end{frontmatter}

\section{Introduction}
\label{sec:intro}

In recent years, the accumulation of data has accelerated across various fields, driven by advancements in computer technologies, widespread use of mobile devices such as smartphones, and development of the Internet of Things, which connects physical objects to the Internet \citep{zhang2022federated,ezugwu2022comprehensive}. In addition, the importance of using such data to discover new knowledge has been increasingly emphasized. Clustering has been widely employed as a representative approach for this purpose \citep{oyewole2023data,singh2024comprehensive}. Clustering is a type of unsupervised learning that partitions a set of data points into distinct groups by maximizing intracluster similarity and minimizing intercluster similarity \citep{altilio2019distributed,ros2024deep}. It enables the discovery of latent patterns within datasets and has been applied to knowledge discovery in diverse domains such as manufacturing \citep{cerquitelli2021enhancing}, civil engineering \citep{cheng2009constraint}, healthcare \citep{khanmohammadi2017improved, perez2021cluster}, and image recognition \citep{villalba2015smartphone}.

Large-scale and diverse datasets are essential to achieve high-quality clustering results, which often require aggregating data distributed across multiple institutions. However, in fields such as healthcare and finance, where data involve individuals or corporations, such integration is challenging owing to issues of confidentiality or privacy \citep{huang2024dra}. Therefore, technical approaches are required to enable the clustering of distributed data without compromising privacy.

In contrast to conventional clustering methods, which assume that data are centrally stored in a single location, federated clustering has recently been proposed as an approach to perform clustering on private data distributed across multiple institutions \citep{dennis2021heterogeneity}. Federated clustering is a branch of federated learning (FL) \citep{mcmahan2017communication} and a privacy-preserving framework for distributed data analysis that focuses specifically on clustering tasks. This enables integrated clustering while preserving privacy by sharing statistical or parametric information rather than raw data among participating institutions \citep{kumar2020federated,qiao2023federated}. It should be noted that in some previous studies, federated clustering refers to methods that cluster institutions themselves based on the distribution of the data they hold (e.g., \cite{ghosh2020efficient}). However, in this study, we used the term to refer to methods that perform integrated clustering of data.

The existing federated clustering methods face several challenges. One major issue lies in the limited assumptions made regarding data distribution. \cite{barcena2024federated} identify two typical scenarios for distributed data analysis using federated clustering. The first is horizontal partitioning, in which each institution holds datasets with the same set of features but different data points. The second is vertical partitioning, in which each institution retains different features corresponding to the same set of data points. However, real-world data often exhibit more complex partitioning structures, such as combinations of horizontal and vertical partitions.

\begin{figure}[tbp]
    \centering
    \includegraphics[width=\textwidth]{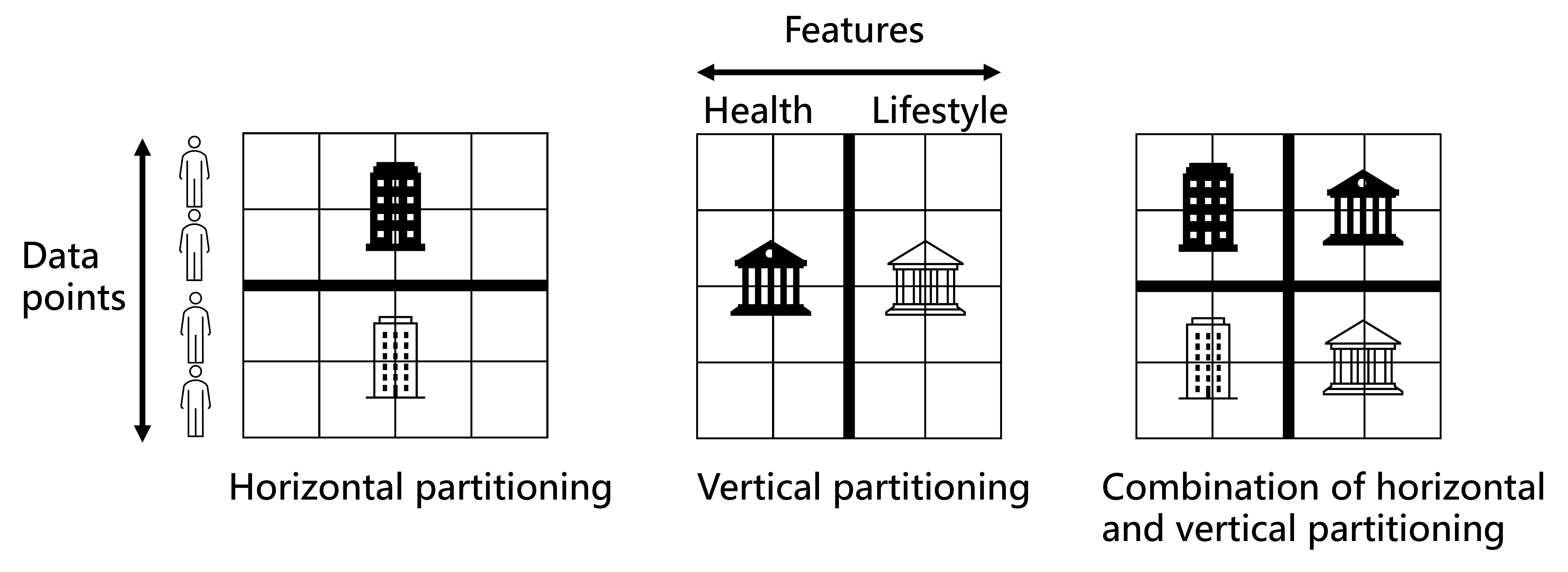}
    \caption{Example of data distribution.}
    \label{fig:data_partition}
\end{figure}

As an example of a scenario involving both types of partitioning, consider a case in which multiple hospitals hold health data for different groups of patients, whereas multiple insurance companies retain lifestyle data for different groups of policyholders (see Fig. \ref{fig:data_partition}). In this situation, vertical partitioning arises because hospitals and insurance companies hold different types of information. Simultaneously, horizontal partitioning is present as each institution handles distinct populations. To perform an integrated analysis that combines information from multiple perspectives, such as health and lifestyle, it is necessary to develop clustering methods that can handle such complex distributed structures.

However, most of the existing federated clustering methods are primarily designed under the assumption of horizontal partitioning, and only a limited number of methods address vertical partitioning. Furthermore, to the best of our knowledge, no federated clustering method has been proposed that can handle scenarios involving a combination of partitioning types.

In this study, we propose a novel federated clustering method called data collaboration clustering (DC-Clustering) to accommodate diverse data distribution scenarios. The proposed method is characterized by its applicability to complex data-partitioning structures, which conventional federated clustering methods cannot handle. Moreover, it offers the flexibility to select either k-means \citep{lloyd1982least} or spectral clustering (SC) \citep{shi2000normalized} depending on the researcher’s objective and enables clustering results to be obtained with only a single round of communication.

The contributions of this study are summarized as follows:
\begin{itemize}
    \item We propose DC-Clustering, a method that enables privacy-preserving integrated clustering.
    \item The proposed method can perform integrated clustering even under complex data distribution scenarios by combining both horizontal and vertical partitioning, which existing methods cannot handle.
    \item This allows for flexible selection between k-means and SC, depending on the researcher's objective.
Each participating local institution can perform integrated clustering using only a single round of communication with a central server (hereafter referred to as the analyst).
    \item Experimental results on both synthetic and open benchmark datasets demonstrate that the proposed method achieves clustering performance comparable to that obtained through centralized clustering, where all data are pooled.
\end{itemize}
The remainder of this paper is organized as follows: Section \ref{sec:relatedwork} reviews the related work, and Section \ref{sec:preliminaries} provides the necessary preliminaries. Section \ref{sec:method} describes the proposed method, DC-Clustering. Section \ref{sec:experiment} presents the experimental settings and the results of evaluating the effectiveness of DC-Clustering. Section \ref{sec:discussion} discusses the experimental results. Finally, Section \ref{sec:conclusion} concludes the study.

\section{Related works}
\label{sec:relatedwork}
This section reviews previous studies related to clustering, data collaboration (DC) analysis, and federated clustering that are closely connected to the proposed approach.

\subsection{Clustering}
\label{sec:clustering}

Clustering is a machine-learning technique that partitions a set of data points into distinct groups. It can be categorized from two perspectives: hierarchical versus non-hierarchical and hard versus soft. Hierarchical clustering builds clusters stepwise using a tree structure. In contrast, non-hierarchical clustering directly partitions data points into mutually independent groups. In hard clustering, each data point is assigned to one cluster, whereas in soft clustering, each data point is assigned probabilistically to multiple clusters.

This study primarily focuses on non-hierarchical and hard clustering methods. Representative examples of these approaches include k-means \citep{lloyd1982least} and SC \citep{shi2000normalized}. k-means is a fundamental method that partitions data points into a predefined number of clusters based on centroids and is widely used because of its computational efficiency. To address the sensitivity of k-means to the initialization of cluster centroids, \citep{arthur2007k} proposed an improved initialization technique called k-means++.

When clusters exhibit complex structures such as non-convex shapes, k-means clustering often fails to produce appropriate partitions. In such cases, SC is considered effective. SC performs spectral embedding by mapping the data points to the eigenspace of a Laplacian matrix constructed from a similarity graph. Clustering is then performed on this embedded representation using standard clustering methods such as k-means clustering.

Clustering algorithms are used to reveal the latent structures and patterns within datasets, making them effective for tasks such as image classification and feature extraction. Clustering has been widely applied as a knowledge discovery method across various domains, including manufacturing, civil engineering, healthcare, and image recognition \citep{choudhary2009data,pan2020mining,lopez2021methodology,oh2022effective,pei2022cluster}.

\subsection{Data Collaboration analysis}
\label{sec:dc_analysis}

In recent years, the importance of performing integrated analysis of distributed data while preserving privacy has been increasingly recognized. FL has been proposed as a framework for addressing this challenge. FL systems can be broadly categorized into model sharing and non-model-sharing types. In model-sharing FL systems \citep{mcmahan2017communication}, each institution shares only the parameters of locally trained models instead of raw data, thereby enabling the construction of an integrated analysis model.

In contrast, DC analysis \citep{imakura2020data}, a non-model-sharing FL system, constructs an integrated analysis model by having each institution generate a dimensionally reduced intermediate representation of its raw data and share it instead. DC analysis is capable of handling complex data distribution scenarios combining both horizontal and vertical partitioning, and it also allows for the flexible selection of analytical models depending on the task. Moreover, as it does not require synchronization or iterative communication between institutions, it offers the advantage of a relatively low communication cost. Although DC analysis was originally developed for predictive tasks such as classification and regression \citep{imakura2020data,imakura2021interpretable,imakura2023non}, its application has recently expanded to a broader range of analytical tasks, including causal inference \citep{kawamata2024collaborative,nakayama2025data}, survival analysis \citep{imakura2023dc,toyoda2025estimating}, feature selection \citep{ye2019distributed}, and anomaly detection \citep{mashiko2025anomaly}.

\subsection{Federated Clustering}
\label{sec:federated_clustering}

Recently, a growing body of research has extended the framework of FL to clustering tasks, known as federated clustering. A pioneering work by \cite{kumar2020federated} proposed federated k-means, which shares cluster centroids as model parameters, and demonstrated its effectiveness on public datasets. Their results suggest that utilizing distributed data can improve clustering performance. Other federated clustering methods that support k-means clustering have also been developed. \cite{dennis2021heterogeneity} proposed a method that operates with a single round of communication and theoretically showed that when each institution maintains distinct clusters, the separation conditions for clustering can be relaxed. \cite{huang2022coresets} explicitly addressed vertical partitioning in the context of federated clustering for the first time. They demonstrated that high-performance clustering results could be achieved while significantly reducing communication costs using a small representative dataset that approximates the entire data distribution.

Federated clustering methods that are compatible with SC have also been proposed. \cite{hernandez2021federated} introduced the first approach to extend the federated clustering framework to SC. Their method enabled each client to construct a local eigenspace while achieving a global cluster structure with minimal communication. Similarly, \cite{qiao2023federated} proposed a federated SC method that reconstructs a similarity matrix without sharing raw data using a federated kernelized factorization approach. Through accuracy evaluations on synthetic datasets, \cite{qiao2023federated} demonstrated that their method achieved clustering performance comparable to that of centralized clustering and outperformed the method of \cite{hernandez2021federated}.

On the other hand, federated clustering methods supporting clustering algorithms other than k-means and SC have also been proposed.
\footnote{In this study, we refer to c-means used for hard clustering as k-means, and explicitly denote its soft clustering counterpart as fuzzy c-means.}
\cite{stallmann2022framework} proposed a method based on fuzzy c-means and constructed a comprehensive framework that included cluster number estimation and validity evaluation. Their results demonstrated that the proposed method achieved stable clustering performance, even when data distributions differed across institutions. \cite{wang2023federated} proposed a federated fuzzy c-means method and compared two approaches, namely model and gradient averaging. Their experiments showed that the gradient averaging approach performed better under a heterogeneous data distribution among institutions. \cite{barcena2024federated} proposed an algorithm capable of performing both k-means and fuzzy c-means, which theoretically yielded results identical to those of centralized clustering and showed high agreement with experimental evaluations.

Among the aforementioned studies, most support only horizontal partitioning, and only a few are applicable to vertical partitioning. Moreover, no existing methods can handle scenarios in which both types of partitions are combined. Bridging this gap is essential to enhance the feasibility of collaborative data analysis across real-world institutions. Table \ref{tab:methods} summarizes the relationship between the existing studies and this study.

\begin{table}[tbp]
    \centering
    \begin{tabular}{ll}
    \hline
    Data distribution setting                                                                      & Available methods                                                                                                                                                                                                                                   \\ \hline
    Horizontal partitioning                                                                        & \begin{tabular}[c]{@{}l@{}}\cite{kumar2020federated}, \cite{dennis2021heterogeneity}, \\ \cite{hernandez2021federated}, \cite{stallmann2022framework}, \\ \cite{qiao2023federated},  \cite{wang2023federated}, \cite{barcena2024federated}, \\ \textbf{This study}\end{tabular} \\ \hline
    Vertical partitioning                                                                          & \begin{tabular}[c]{@{}l@{}}\cite{huang2022coresets}, \cite{barcena2024federated},\\ \textbf{This study}\end{tabular}                                                                                                                                                \\ \hline
    \begin{tabular}[c]{@{}l@{}}Combination of horizontal \\ and vertical partitioning\end{tabular} & \textbf{This study} \\ \hline
    \end{tabular}
    \caption{Positioning of this study in comparison with existing federated clustering studies.}
    \label{tab:methods}
\end{table}

In addition, the characteristics of the aforementioned studies can be organized from the perspectives of clustering algorithm flexibility and communication efficiency. Currently, only the method proposed by \cite{barcena2024federated} allows for the selection of multiple clustering algorithms. However, the two algorithms supported by these methods, c-means and fuzzy c-means, are based on relatively simple data structures. By contrast, designs that allow for the complementary use of structurally different clustering algorithms, such as k-means and SC, may offer greater flexibility in handling more complex data structures. 

Regarding communication efficiency, \cite{dennis2021heterogeneity} and \cite{hernandez2021federated} proposed methods that can perform clustering with only a single round of communication. Such high communication efficiency is particularly valuable in environments with restricted network connectivity, for example, in scenarios involving medical data stored on isolated systems.

\section{Preliminaries}
\label{sec:preliminaries}

This section introduces the preliminaries of the proposed method. Section \ref{sec:clustering_algorithm} describes the clustering algorithms, and Section \ref{sec:distributed_data_setting} explains the data-partitioning settings. Throughout this section, we consider a dataset $X=[x_1, \cdots , x_n]^{\top} \in \mathbb{R}^{n \times m}$, where $n$ is the sample size, $m$ is the number of features, and $x_i \in \mathbb{R}^m$ represents the feature vector of the $i$th data point.

\subsection{Clustering Algorithm}
\label{sec:clustering_algorithm}

This section provides an overview of the clustering algorithms used in the study. Section \ref{sec:k-means_clustering} describes k-means, and Section \ref{sec:spectral_clustering} explains SC.

\subsubsection{k-means Clustering}
\label{sec:k-means_clustering}

k-means is a clustering algorithm that partitions data points into k clusters such that the average squared distance between data points within the same cluster is minimized. In this study, we focus on k-means++, an improved version of k-means that addresses the sensitivity to the initial selection of cluster centroids. Hereafter, we refer to k-means++ simply as k-means. The algorithm is outlined in Algorithm \ref{alg:kmeans} and consists of two steps.
\\
\textit{Step 1: Initialization of cluster centroids}\\
In this step, one data point is randomly selected from the dataset X and assigned as the first cluster centroid $\theta_1^{\text{KM}} \in \mathbb{R}^m$. For each of the remaining centroids $\theta_j^{\text{KM}} \in \mathbb{R}^m  (j=2, \cdots ,k)$ the following procedure is applied sequentially. First, for each data point $i$, the distance $D_i$ to its nearest cluster centroid among those already selected is computed. Then, the next centroid $\theta_j^{KM}$ is chosen randomly according to the probability distribution defined by (\ref{eq:kmeans_centroid}).
\begin{equation}
    \text{Pr}(\theta_j^{\text{KM}})=\frac{D_i^2}{\sum_{i^\prime = 1}^n D_{i^\prime}^2} ~ (i=1, \cdots, n).
    \label{eq:kmeans_centroid}
\end{equation}
\\
\textit{Step 2: Update of cluster centroids}\\
In this step, the following procedure is repeated for $R$ iterations: First, for each data point $i$, the nearest cluster $A_i$ is identified based on the current centroids. Then, update the centroid $\theta_{A_i}^{\text{KM}}$ by computing the average of all $x_i$ that belong to the same cluster $A_i$. After the iterations are complete, the final cluster centroids $\theta_1^\text{KM},\cdots,\theta_k^\text{KM}$ and the estimated cluster labels $\hat{Y}=(A_1,\cdots ,A_n)$ are obtained.

\begin{algorithm}[tb]
    \caption{k-means}
    \label{alg:kmeans}
    \begin{algorithmic}[1]
        \Statex \textbf{Input: Dataset $X$, Number of clusters $k$, Number of iterations $R$}.
        \Statex \textbf{Output: Estimated cluster labels $\widehat{Y}$, Cluster centers $\theta_1^{\text{KM}}, \dots, \theta_k^{\text{KM}}$}.
        \vspace{-.3\baselineskip}
        
        \Statex \hrulefill
        \State Get the cluster center $\theta_1^{\text{KM}}$ from $X$ randomly.
        \State \textbf{for} $j = 2, \dots, k$ : \textbf{do}
        \State \qquad \textbf{for} $i = 1, \dots, n$ : \textbf{do}
        \State \qquad \qquad Compute $D_i$, the distance to $x_i$ from the nearest cluster center of $x_i$.
        \State \qquad \textbf{end for}
        \State \qquad Get the cluster center $\theta_j^{\text{KM}}$ according to (1).
        \State \textbf{end for}
        \State \textbf{for} $l = 2, \dots, R$ : \textbf{do}
        \State \qquad \textbf{for} $i = 1, \dots, n$ : \textbf{do}
        \State \qquad \qquad Find $A_i$, the nearest cluster center of $x_i$.
        \State \qquad \textbf{end for}
        \State \qquad \textbf{for} $j = 1, \dots, k$ : \textbf{do}
        \State \qquad \qquad $\theta_j^{\text{KM}} \leftarrow \text{mean}\{x_i \mid A_i = j\}$
        \State \qquad \textbf{end for}
        \State \textbf{end for}
        \State Return $\widehat{Y}$ and $\theta_1^{\text{KM}}, \dots, \theta_k^{\text{KM}}$.
    \end{algorithmic}
\end{algorithm}

\subsubsection{Spectral Clustering (SC)}
\label{sec:spectral_clustering}

SC first constructs a graph from the dataset, where each data point is treated as a node and the similarity between data points is represented by weighted edges. The constructed graph is then partitioned into k subgraphs such that the total weight of the edges being cut is minimized. \cite{luxburg2007tutorial} showed that this graph-partitioning problem can be approximated by solving the eigenvalue problem of a graph Laplacian matrix. The dataset is then transformed into a new representation called spectral embedding. Finally, a clustering algorithm is applied to this transformed representation. The overall SC procedure is summarized in Algorithm \ref{alg:spectral} and consists of the following four steps:
\\
\textit{Step 1: Construction of affinity matrix}\\
In this step, an affinity matrix $W \in \mathbb{R}^{n \times n}$ is constructed based on the dataset $X$. 
The affinity matrix stores the pairwise similarities between data points in the dataset, where element $w_{i,j}$ represents the similarity between $x_i$ and $x_j$. In this study, we adopted the k-nearest neighbor approach, which is a common method for constructing affinity matrices. 
In this method, if $x_j$ is among the $k$ nearest neighbors of $x_i$, the corresponding element $w_{i,j}$ is set to 1; otherwise, it is set to 0.\\
\textit{Step 2: Construction of Laplacian matrix}\\
In this step, the Laplacian matrix $L \in \mathbb{R}^{n \times n}$ is constructed based on the previously computed affinity matrix. The Laplacian matrix is calculated using (\ref{eq:spectral_l}).
\begin{equation}
    L=D-W.
    \label{eq:spectral_l}
\end{equation}
where $D$ is the degree matrix. In this study, a spectral embedding representation is constructed from the normalized Laplacian matrix $L_\text{sym}$, as defined in (\ref{eq:spctral_lsym}).
\begin{equation}
    L_\text{sym} = D^{-\frac{1}{2}} L D^{-\frac{1}{2}} = I - D^{-\frac{1}{2}} W D^{-\frac{1}{2}}.
    \label{eq:spctral_lsym}
\end{equation}
\textit{Step 3: Construction of spectral embedding representation}\\
In this step, $k$ eigenvectors $e_1, \cdots ,e_k \in \mathbb{R}^n$ of the Laplacian matrix are extracted in ascending order of their corresponding eigenvalues. These eigenvectors are then assembled to form the matrix $E=[e_1, \cdots,e_k ] \in \mathbb{R}^{n \times k}$, which is referred to as the spectral embedding representation.
\\
\textit{Step 4: Clustering}\\
In this step, a standard clustering algorithm is applied to matrix E to partition the data into $k$ clusters. In this study, we adopt the k-means algorithm to cluster E. As a result of applying k-means, we obtain the cluster centroids $\theta_1^\text{SC}, \cdots ,\theta_k^\text{SC} \in \mathbb{R}^k$, along with the estimated cluster labels $\hat{Y}=(\hat{y}_1,\cdots,\hat{y}_n)$.

\begin{algorithm}[tb]
    \caption{Spectral clustering}
    \label{alg:spectral}
    \begin{algorithmic}[1]
        \Statex \textbf{Input: Dataset $X$, Number of clusters $k$}.
        \Statex \textbf{Output: Estimated cluster labels $\widehat{Y}$, Cluster centers $\theta_1^{\text{SC}}, \dots, \theta_k^{\text{SC}}$}.
        \vspace{-.3\baselineskip}
        
        \Statex \hrulefill
        \State Get the affinity matrix $W$ from the dataset $X$.
        \State Compute the Laplacian matrix $L$ from $W$.
        \State Normalize $L$ to get $L_{\text{sym}}$.
        \State Compute the eigenvectors $e_1, \dots, e_k$ from $L_{\text{sym}}$.
        \State Get the spectral embedding $E = [e_1, \dots, e_k]$.
        \State Do k-means on $E$ and get $\widehat{Y}$ and $\theta_1^{\text{SC}}, \dots, \theta_k^{\text{SC}}$.
        \State Return $\widehat{Y}$ and $\theta_1^{\text{SC}}, \dots, \theta_k^{\text{SC}}$.
        
    \end{algorithmic}
\end{algorithm}

\subsection{Distributed Data Setting}
\label{sec:distributed_data_setting}
In this study, we considered two types of data-partitioning settings: horizontal and vertical. Horizontal partitioning refers to a scenario in which different institutions hold data points sharing the same feature space. By contrast, vertical partitioning refers to a scenario in which different institutions hold different sets of features for the same data points.

In this study, we assumed a setting in which dataset $X$ was partitioned into $c$ rows along the data point axis and $d$ columns along the feature axis in a lattice pattern, resulting in a total of $c \times d$ participating institutions. Let $n_i$ and $m_j$ denote the number of data points and features held by an institution located in the $i$th row and $j$th column of this grid, respectively. The distributed data $X_{i,j}$ are then represented as shown in (4).
\begin{equation}
    X = \begin{bmatrix}
        X_{1,1} & X_{1,2} & \cdots & X_{1,d} \\
        X_{2,1} & X_{2,2} & \cdots & X_{2,d} \\
        \vdots & \vdots & \ddots & \vdots \\
        X_{c,1} & X_{c,2} & \cdots & X_{c,d}
    \end{bmatrix} \in \mathbb{R}^{n \times m}, \quad X_{i,j} \in \mathbb{R}^{n_i \times m_j}.
\end{equation}
When $c>1$, the data distribution can be interpreted as horizontal partitioning, and when $d>1$, it corresponds to vertical partitioning. In cases where both $c>1$ and $d>1$, the dataset is considered to be distributed in a complex manner that combines horizontal and vertical partitions, as described in the Introduction. Such complex settings cannot be handled using existing methods. 
Note that \cite{imakura2021collaborative} states that DC analysis is applicable to non-lattice-like partitioned data. While our method can also be applied to non-lattice-like partitioned data, this paper does not address such partitions.
Let $(i,j)$ denote the institutions located in the $i$th row and $j$th column of the partitioned data matrix, respectively.

\section{Method}
\label{sec:method}

We first present the proposed method in Section \ref{sec:data_collaboration_clustering_framework}. Then, in Section \ref{sec:discussion_on_privacy-preserving_of_DC-Clustering}, we discuss the privacy-preserving aspects of the proposed method. Finally, Section \ref{sec:functions_for_intermediate_and_collaborative_representations} addresses the construction of collaborative representation in the proposed framework.

\subsection{Data Collaboration Clustering Framework}
\label{sec:data_collaboration_clustering_framework}
We propose DC-Clustering, an extension of DC analysis \citep{imakura2020data}, as a framework that enables the clustering of distributed data while preserving privacy. DC-Clustering’s procedure shares a commonality with other DC analysis methods in that it constructs a collaborative representation. However, it differs in two key aspects: (1) the collaborative representation is utilized specifically for clustering and (2) the information returned to each participating institution consists of clustering representations and cluster centroids. Similar to other DC analysis methods, the proposed framework involves two roles: the user and the analyst.

\begin{figure}[tbp]
    \centering
    \includegraphics[width=0.5\textwidth]{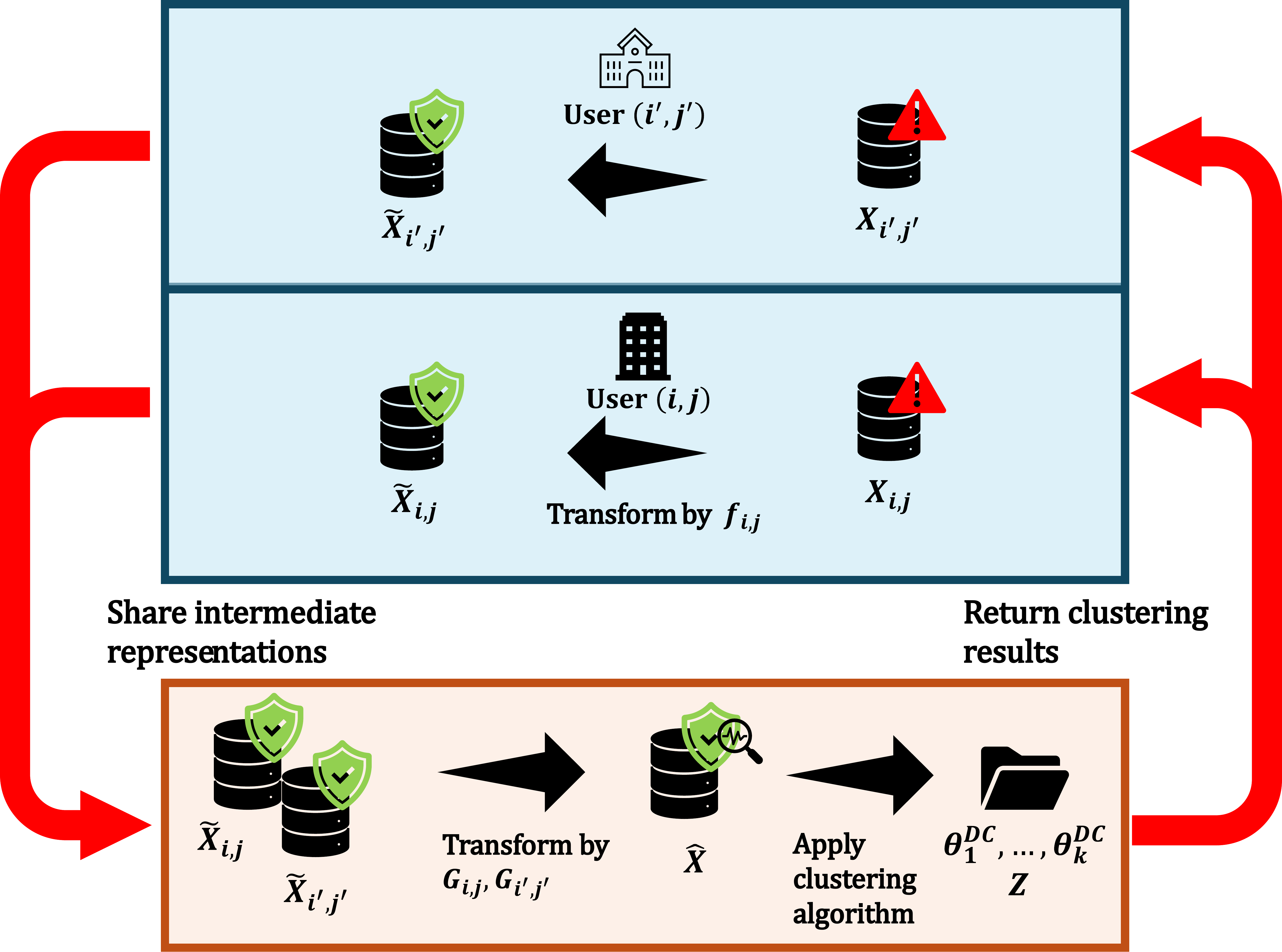}
    \caption{Overview of proposed method.}
    \label{fig:method}
\end{figure}

The proposed method consists of two main stages: (1) construction of a collaborative representation, and (2) execution and sharing of clustering, as illustrated in Fig. \ref{fig:method}. In the first stage, each user constructs an intermediate representation independently and shares it with the analyst, who then builds a collaborative representation. In the second stage, the analyst performs clustering based on the collaborative representation and sends the results back to the users, allowing them to obtain clustering outcomes for their own datasets. The algorithm is outlined in Algorithm \ref{alg:dc_clustering}. The details of the first and second stages are presented in Sections \ref{sec:construction_of_collaborative_representations} and \ref{sec:clustering_and_result_sharing}, respectively.

\begin{algorithm}[tb]
    \caption{Data collaboration clustering}
    \label{alg:dc_clustering}
    \begin{algorithmic}[1]
        \Statex \textbf{Input: $X_{i,j}$: Dataset of user $(i,j)$, $k$: Number of clusters}.
        \Statex \textbf{Output: $\widehat{Y}, \widehat{Y}_1, \dots, \widehat{Y}_c$: Estimated cluster labels for each user}.
        \vspace{-.3\baselineskip}
        
        \Statex \hrulefill
        \Statex \textbf{User$(i,j)$ side \hspace{0.22\linewidth} Analyst side}
        
        \State Get anchor dataset $X^{\text{anc}}$ and set $X_{:,j}^{\text{anc}}$.
        \State Generate $f_{i,j}$.
        \State Compute $\widetilde{X}_{i,j} = f_{i,j}(X_{i,j})$.
        \State Compute $\widetilde{X}_{i,j}^{\text{anc}} = f_{i,j}(X_{:,j}^{\text{anc}})$.
        
        \State \begin{tabular}[t]{@{}p{0.45\linewidth}@{\hspace{1em}}p{0.05\linewidth}@{}p{0.45\linewidth}@{}} Share $\widetilde{X}_{i,j}$ and $\widetilde{X}_{i,j}^{\text{anc}}$. & $\searrow$ & \end{tabular}
        
        \State \begin{tabular}[t]{@{}p{0.45\linewidth}@{\hspace{1em}}p{0.05\linewidth}@{}p{0.45\linewidth}@{}} & & Get $\widetilde{X}_{i,j}$ and $\widetilde{X}_{i,j}^{\text{anc}}$ from all users. \end{tabular}
        \State \begin{tabular}[t]{@{}p{0.45\linewidth}@{\hspace{1em}}p{0.05\linewidth}@{}p{0.45\linewidth}@{}} & & Set $\widetilde{X}_i$ and $\widetilde{X}_i^{\text{anc}}$ for all $i$. \end{tabular}
        \State \begin{tabular}[t]{@{}p{0.45\linewidth}@{\hspace{1em}}p{0.05\linewidth}@{}p{0.45\linewidth}@{}} & & Compute $g_i$ satisfying (10) for all $i$. \end{tabular}
        \State \begin{tabular}[t]{@{}p{0.45\linewidth}@{\hspace{1em}}p{0.05\linewidth}@{}p{0.45\linewidth}@{}} & & Compute $\widehat{X}_i = g_i(\widetilde{X}_i)$ for all $i$. \end{tabular}
        \State \begin{tabular}[t]{@{}p{0.45\linewidth}@{\hspace{1em}}p{0.05\linewidth}@{}p{0.45\linewidth}@{}} & & Set $\widehat{X}$. \end{tabular}
        \State \begin{tabular}[t]{@{}p{0.45\linewidth}@{\hspace{1em}}p{0.05\linewidth}@{}p{0.45\linewidth}@{}} & & Do k-means or SC on $\widehat{X}$, \\ & & then get $\theta_1^{\text{DC}}, \dots, \theta_k^{\text{DC}}$ and $Z$. \end{tabular}
        \State \begin{tabular}[t]{@{}p{0.45\linewidth}@{\hspace{1em}}p{0.05\linewidth}@{}p{0.45\linewidth}@{}} & & Split $Z$ into $Z_1, \dots, Z_c$. \end{tabular}
        
        \State \begin{tabular}[t]{@{}p{0.45\linewidth}@{\hspace{1em}}p{0.05\linewidth}@{}p{0.45\linewidth}@{}} & $\swarrow$ & Share $\theta_1^{\text{DC}}, \dots, \theta_k^{\text{DC}}$ and $Z_i$ with $i$th row users. \end{tabular}
        
        \State Get $\theta_1^{\text{DC}}, \dots, \theta_k^{\text{DC}}$ and $Z_i$.
        \State Compute $\widehat{Y}_i$ from $\theta_1^{\text{DC}}, \dots, \theta_k^{\text{DC}}$ and $Z_i$.
        
    \end{algorithmic}
\end{algorithm}

\subsubsection{Construction of collaborative representations}
\label{sec:construction_of_collaborative_representations}

In the first stage of the proposed method, all users share an anchor dataset $X^\text{anc} \in R^{r \times m}$. The anchor dataset $X^\text{anc}$ is a publicly available or randomly generated dataset that can be shared among institutions, where $r$ denotes the sample size of the anchor dataset. An anchor dataset is used to compute the transformation functions for constructing a collaborative representation. Various methods exist for generating anchor datasets, including uniform sampling, low-rank approximation, and oversampling \citep{imakura2023another}.

Next, each user $(i,j)$ constructs an intermediate representation $\tilde{X}_{i,j}$ of their local dataset $X_{i,j}$ as well as $\tilde{X}_{:,j}^\text{anc}$, based on the anchor dataset $X_{:,j}^\text{anc}$, using (\ref{eq:inter_rep_raw_ij}) and (\ref{eq:inter_rep_anc_ij}).
\begin{align}
    \tilde{X}_{i,j} &= f_{i,j}(X_{i,j}) \in \mathbb{R}^{n_i \times \tilde{m}_{i,j}}, \label{eq:inter_rep_raw_ij} \\
    \tilde{X}_{i,j}^{\text{anc}} &= f_{i,j}(X_{i,j}^{\text{anc}}) \in \mathbb{R}^{r \times \tilde{m}_{i,j}}, \label{eq:inter_rep_anc_ij}
\end{align}
where $f_{i,j}$ is a function that generates an intermediate representation by performing row-wise dimensionality reduction. Techniques such as principal component analysis (PCA) \citep{pearson1901liii}, locality-preserving projection \citep{he2003locality}, and autoencoders \citep{hinton2006reducing} can be used to implement this function. Note that $f_{i,j}$ is a private function that is not shared with other users. The variables $\tilde{m}_{i,j} (<m_{i,j})$ denote the dimensionality of the intermediate representation. $X_{:,j}^\text{anc}$ refers to the anchor dataset restricted to the features held by users in column $j$. The intermediate representations $\tilde{X}_{i,j}$ and $\tilde{X}_{i,j}^\text{anc}$ obtained from the above equations are shared with the analyst instead of the original data $X_{i,j}$. On the analyst's side, the intermediate representations received are arranged as defined in (\ref{eq:inter_rep_raw}) and (\ref{eq:inter_rep_anc}).
\begin{align}
    \tilde{X}_i &= [\tilde{X}_{i,1}, \tilde{X}_{i,2}, \dots, \tilde{X}_{i,d}] \in \mathbb{R}^{n_i \times \tilde{m}_i}, \label{eq:inter_rep_raw} \\
    \tilde{X}_i^{\text{anc}} &= [\tilde{X}_{i,1}^{\text{anc}}, \tilde{X}_{i,2}^{\text{anc}}, \dots, \tilde{X}_{i,d}^{\text{anc}}] \in \mathbb{R}^{r \times \tilde{m}_i}, \label{eq:inter_rep_anc}
\end{align}
The dimensionality $\tilde{m}_i$ of the intermediate representation is defined as shown in (\ref{eq:mtsum}).
\begin{equation}
    \tilde{m}_i = \sum_{j=1}^{d} \tilde{m}_{i,j} \label{eq:mtsum}
\end{equation}

As $f_{i,j}$ may differ across users, the intermediate representations obtained from different users are not directly comparable. Therefore, the analyst cannot simply concatenate the shared intermediate representations for clustering. Instead, the analyst employs a collaborative representation that transforms the intermediate representations into a comparable format. To this end, the analyst constructs a transformation function $g_i$ that satisfies (\ref{eq:g_approx}).
\begin{equation}
    g_i(\tilde{X}_i^{\text{anc}}) \approx g_{i'}(\tilde{X}_{i'}^{\text{anc}}) \in \mathbb{R}^{r \times \widehat{m}} \quad (i \neq i') . \label{eq:g_approx}
\end{equation}
The transformation function $g_i$ can be derived using methods such as an algorithm based on singular value decomposition (SVD), as described in Section 4.3. Based on the obtained function $g_i$, the collaborative representation is given as shown in (\ref{eq:x_hat}).
\begin{equation}
    \widehat{X} = 
    \begin{bmatrix}
    \widehat{X}_1 \\
    \vdots \\
    \widehat{X}_c
    \end{bmatrix} = 
    \begin{bmatrix}
    g_1(\tilde{X}_1) \\
    \vdots \\
    g_c(\tilde{X}_c)
    \end{bmatrix} \in \mathbb{R}^{n \times \widehat{m}} . \label{eq:x_hat}
\end{equation}

\subsubsection{Clustering and result sharing}
\label{sec:clustering_and_result_sharing}
In the second stage of the proposed method, either k-means clustering or SC can be employed. Here, we define the following clustering representation $Z$ in (\ref{eq:z}).
\begin{equation}
    Z = \begin{cases}
    \widehat{X} & (\text{k-means}) \\
    E & (\text{SC})
    \end{cases}, \label{eq:z}
\end{equation}
where $E$ is the spectral embedding constructed from $\hat{X}$. By applying k-means to $Z$, we obtain the cluster centroids $\theta_1^\text{DC}, \cdots, \theta_k^\text{DC}$. After the clustering is performed, the analyst shares $\theta_1^\text{DC}, \cdots, \theta_k^\text{DC}$ and $Z_i$ with each institution $(i,j)$, where $Z_i$ is the clustering representation corresponding to the dataset of the ith row users. Then, user $(i,j)$ determines the nearest cluster for each data point in $Z_i$ by comparing it with the centroids $\theta_1^\text{DC}, \cdots,\theta_k^\text{DC}$, and obtains the estimated cluster labels $\hat{Y}_i$ for their own dataset $X_{i,j}$.

\subsection{Discussion on Privacy-preserving of DC-Clustering}
\label{sec:discussion_on_privacy-preserving_of_DC-Clustering}
This section discusses the privacy-preserving properties of the proposed method, DC-Clustering. Similar to existing data collaboration analysis methods \citep{imakura2021interpretable}, the proposed method incorporates a two-layer privacy protection scheme to ensure that the original dataset $X_{i,j}$ cannot be estimated from the intermediate representation $\tilde{X}_{i,j}$.

In the first layer, no third party possesses both the input and output of transformation function $f_{i,j}$. Consequently, it is impossible for others to estimate the original dataset $X_{i,j}$ from the intermediate representation $\tilde{X}_{i,j}$, even by approximating the inverse of $f_{i,j}$. The second layer is that $f_{i,j}$ is the dimensionality-reduction function. Even if $f_{i,j}$ is leaked, only an approximate inverse can be derived from it; therefore, a perfect reconstruction of $X_{i,j}$ from $\tilde{X}_{i,j}$ is not possible. This two-layered privacy protection mechanism ensures the privacy of DC-Clustering.

Furthermore, we discuss that, as long as the procedure described in Algorithm 3 is followed, user $(i,j)$ cannot fully reconstruct or estimate user $(i,j^\prime )$'s $(j\neq j^\prime )$ dataset $X_{i,j}$ through the shared clustering representation $Z_i$. First, it is evident that a perfect reconstruction is impossible because $f_{i,j}$ is a dimensionality-reduction function. Next, we consider the infeasibility of the estimation. The representation $Z_i$ is constructed based on the intermediate representation obtained by applying $f_{i,j}$ to $X_{i,j}$. Therefore, to estimate $X_{i,j}$ from $Z_i$, one must at least derive an approximate inverse of $f_{i,j}$. However, no other party possesses either the input or the output of $f_{i,j}$, which makes such an approximation infeasible. Hence, user $(i,j^\prime )$ cannot estimate $X_{i,j}$ from $Z_i$.

Within the framework of DC analysis, various types of attacks aimed at reconstructing dataset $X_{i,j}$ can be envisioned. For example, \cite{imakura2021accuracy} demonstrated that $X_{i,j}$ may leak to other parties through collusion among users. Countermeasures against such attacks remain important subjects for future research.

\subsection{Functions for intermediate and collaborative representations}
\label{sec:functions_for_intermediate_and_collaborative_representations}
This section provides a detailed explanation of the intermediate representation function $f_{i,j}$ and derivation of the collaborative representation function $g_i$ in the proposed method. Here, we consider a dimensionality reduction based on affine transformations, such as PCA, in the form of an intermediate representation function. An affine transformation is a type of transformation that combines a linear transformation with a translation and can be used to express a process in which data are first centered and then linearly transformed. The intermediate representation $\tilde{X}_{i,j}$ is expressed as (\ref{eq:inter_rep_raw_ij_affine}).
\begin{equation}
    \tilde{X}_{i,j} = f_{i,j}(X_{i,j}) = (X_{i,j} - \mathbf{1} \boldsymbol{\mu}_{i,j}) F_{i,j}, \label{eq:inter_rep_raw_ij_affine}
\end{equation}
where $\mathbf{1}=[1,\cdots,1]^\top$, and $\boldsymbol{\mu}_{i,j} \in \mathbb{R}^{1 \times m_{i,j} }$ is a vector composed of the average values when centering is applied, or zeros otherwise. $F_{i,j} \in \mathbb{R}^{n_i \times \tilde{m}_{i,j} }$ denotes a linear transformation. Let
\begin{equation}
    F_i = 
    \begin{bmatrix}
    F_{i,1} &  &  \\
     & \ddots &  \\
     &  & F_{i,d}
    \end{bmatrix} , \quad
    \boldsymbol{\mu}_i = [\boldsymbol{\mu}_{i,1}, \dots, \boldsymbol{\mu}_{i,d}] . \label{eq:f_mu}
\end{equation}
Then, the intermediate representation is $\tilde{X}_i=([X_{i,1}, \cdots, X_{i,d} ]- \mathbf{1} \mu_i ) F_i$. Similarly, the intermediate representation of the anchor data is $\tilde{X}_i^\text{anc}=([X_{i,1}^\text{anc},\cdots,X_{i,d}^\text{anc} ]-\mathbf{1} \mu_i ) F_i$.

When the collaborative representation function is assumed to be a linear transformation $G_i \in \mathbb{R}^{\tilde{m}_i \times \hat{m}}$, an approximate derivation method for this transformation was proposed by \cite{imakura2020data}. In this case, the function is expressed as $g_i (\tilde{X}_i^\text{anc})=\tilde{X}_i^\text{anc} G_i$. The matrix $G_i$ can be derived by applying SVD to $\tilde{X}_i^\text{anc}$ as shown in (\ref{eq:svd}).
\begin{equation}
    [\tilde{X}_1^{\text{anc}}, \tilde{X}_2^{\text{anc}}, \dots, \tilde{X}_c^{\text{anc}}] = [U_1 \ U_2] 
    \begin{bmatrix}
    \Sigma_1 &  \\
     & \Sigma_2
    \end{bmatrix}
    \begin{bmatrix}
    V_1^{\top} \\
    V_2^{\top}
    \end{bmatrix} \approx U_1 \Sigma_1 V_1^{\top}, \label{eq:svd}
\end{equation}
where $\Sigma_1 \in \mathbb{R}^{\hat{m} \times \hat{m}}$ is a diagonal matrix whose diagonal elements capture the majority of the singular values, and $U_1$ and $V_1$ are orthogonal matrices whose columns consist of the corresponding left and right singular vectors, respectively. Matrix $G_i$ is computed using (\ref{eq:g}).
\begin{equation}
    G_i = (\tilde{X}_i^{\text{anc}})^{\dagger} U_1 . \label{eq:g}
\end{equation}
The symbol $\dagger$ denotes the Moore–Penrose pseudo-inverse. Accordingly, the collaborative representation was obtained as $\hat{X}_i= \tilde{X}_i G_i$.

Many previous studies on DC-related methods assumed that the collaborative representation function is a linear transformation. However, when an intermediate representation function is an affine transformation such as PCA, it is intuitively reasonable to assume that the collaborative representation function should also be affine, as this may yield better collaborative representations. A theoretical justification for this intuition is provided in \ref{app:affine}, and in this section, we describe how to derive the function $g_i$ under the affine transformation assumption. When $g_i$ is assumed to be an affine transformation, it can be derived in approximately the same manner as that in the linear case. In this case, the collaborative representation of $\tilde{X}_i^\text{anc}$ can be expressed as
\begin{equation}
    g_i(\tilde{X}_i^{\text{anc}}) = [\tilde{X}_i^{\text{anc}}, \mathbf{1}] 
    \begin{bmatrix}
    G_i' \\
    \mathbf{G}_i^*
    \end{bmatrix} = \tilde{X}_i^{\text{anc}} G_i' + \mathbf{1} \mathbf{G}_i^*,
\end{equation}
where $G_i^\prime \in \mathbb{R}^{\tilde{m}_i \times \hat{m}}, G_i^* \in \mathbb{R}^{1 \times \hat{m}}$. The pair $(G_i^\prime, G_i^* )$ defines the affine transformation $g_i$. Matrices $G_i^\prime$ and $G_i^*$ can be derived by applying SVD to matrix $[\tilde{X}_i^\text{anc},\mathbf{1}]$ as shown in (\ref{eq:svd_affine}).
\begin{equation}
    [\tilde{X}_1^{\text{anc}}, \mathbf{1}, \tilde{X}_2^{\text{anc}}, \mathbf{1}, \dots, \tilde{X}_c^{\text{anc}}, \mathbf{1}] \approx \bar{U}_1 \bar{\Sigma}_1 \bar{V}_1^\top , \label{eq:svd_affine}
\end{equation}
where $\bar{U}_1$, $\bar{\Sigma}_1$ and $\bar{V}_1$ are defined in the same manner as in (\ref{eq:svd}). Based on this, the collaborative representation function and collaborative representation are given by (\ref{eq:g_affine}) and (\ref{eq:x_hat_affine}), respectively.
\begin{align}
    \begin{bmatrix} G_i\prime \\ \mathbf{G}_i^* \end{bmatrix} &= [\tilde{X}_i^{\text{anc}}, \mathbf{1}]^{\dagger} \bar{U}_1, \label{eq:g_affine} \\
    \widehat{X}_i &= [\tilde{X}_i, \mathbf{1}] \begin{bmatrix} G_i^\prime \\ \mathbf{G}_i^* \end{bmatrix}. \label{eq:x_hat_affine}
\end{align}

\section{Experiment}
\label{sec:experiment}
In this section, the performance of the proposed method is evaluated through numerical experiments. Section \ref{sec:common_setting_and_evaluation_scheme} describes the common experimental settings and evaluation scheme. In Section \ref{sec:experiment_I_proof-of-concept_using_synthetic_data}, we describe a proof-of-concept experiment using synthetic data to confirm the effectiveness of the proposed method. In Section \ref{sec:experiment_II_evaluation_performance_using_open_data}, experiments are performed using open benchmark datasets to examine the applicability of the proposed method in real-world scenarios. The common objective of both experiments was to verify whether the proposed method could achieve a clustering performance comparable to that of the ideal case, where all data are centrally aggregated, referred to as centralized clustering (described later).
\footnote{The experiments were run on a macOS Ventura environment computer with an Apple M1 and 8 GB of RAM. Python 3.8.8 and scikit-learn 1.2.2 were used to implement the experimental code. The experimental code is available by reasonable request to the corresponding author.}

\subsection{Common Setting and Evaluation Scheme}
\label{sec:common_setting_and_evaluation_scheme}
We conducted performance comparison experiments focusing on both k-means and SC in Sections \ref{sec:k-means_clustering} and \ref{sec:spectral_clustering}. In the k-means-based experiments, we compared the results of k-means clustering applied in three settings: DC-Clustering, local clustering, and centralized clustering. Local clustering refers to clustering performed using only the local dataset $X_{i,j}$ held by an individual institution. By contrast, centralized clustering refers to clustering performed using the entire dataset $X$, which represents the ideal case. Similarly, in the experiments focusing on SC, we compared the results of SC applied to DC-Clustering, local clustering, and centralized clustering.

The experiments were conducted under the following conditions: The anchored dataset $X^\text{anc}$ was generated using random values drawn from a uniform distribution defined over the range between the minimum and maximum values of each feature in $X$, with the sample size set to $r=n$. The intermediate representation function $f_{i,j}$ was implemented as a sequence of transformations consisting of standardization followed by PCA. The dimensionality $\tilde{m}_{i,j}$ of the intermediate representation was set to one less than the number of features in the dataset held by the institution $(i,j)$. The collaborative representation function was assumed to be an affine transformation. This assumption was adopted based on our preliminary numerical experiments, in which the affine assumption yielded better clustering performance than the assumption of a linear transformation. The number of iterations for k-means clustering was set to R=300. The number of clusters k to be explored was matched to the ground-truth number of clusters, as described later.

To evaluate the performance of each method, we used three widely recognized metrics for clustering evaluation: the adjusted Rand index (ARI), normalized mutual information (NMI), and clustering accuracy (ACC). These metrics quantitatively assess the correspondence between the ground-truth and predicted cluster labels. Let the ground-truth cluster labels be denoted by $Y=(y_1,\cdots,y_n)$, and the predicted cluster labels be $\hat{Y}=(\hat{y}_1,\cdots,\hat{y}_n )$. ARI measures the similarity between $Y$ and $\hat{Y}$, and is defined in (\ref{eq:ari}) \citep{hubert1985comparing}:
\begin{equation}
    \text{ARI} = \frac{\sum_{i,j} \binom{u_{i,j}}{2} - \left[ \sum_i \binom{a_i}{2} \sum_j \binom{b_j}{2} \right] / \binom{n}{2}}{\frac{1}{2} \left[ \sum_i \binom{a_i}{2} + \sum_j \binom{b_j}{2} \right] - \left[ \sum_i \binom{a_i}{2} \sum_j \binom{b_j}{2} \right] / \binom{n}{2}}, \label{eq:ari}
\end{equation}
where $u_{i,j}$ denotes the number of data points whose ground-truth label is $i$ and predicted label is $j$, and $a_i=\sum_j u_{i,j}$, $b_j=\sum_i u_{i,j}$. NMI is a metric that compares the mutual information between $Y$ and $\hat{Y}$, normalized by their entropy, and is defined as (\ref{eq:nmi}) \citep{strehl2002cluster}.
\begin{equation}
    \text{NMI}(\widehat{Y}, Y) = \frac{I(\widehat{Y}; Y)}{\sqrt{H(\widehat{Y})H(Y)}}, \label{eq:nmi}
\end{equation}
where, $I(\hat{Y};Y)$ represents the mutual information between $\hat{Y}$ and $Y$, and $H(\cdot)$ denotes the entropy. ACC measures the classification accuracy using the optimal label mapping between $Y$ and $\hat{Y}$, and is defined as (\ref{eq:acc}) \citep{xu2003document}:
\begin{equation}
    \text{ACC} = \max_{\pi \in \mathcal{P}} \frac{1}{n} \sum_{i} 1(y_i = \pi(\widehat{y}_i)). \label{eq:acc}
\end{equation}
Here, $\mathcal{P}$ denotes all possible permutations of the mapping between ground-truth and predicted cluster labels, and $1(\cdot)$ is the indicator function. All three evaluation metrics assign higher values to better clustering performance and reach a maximum value of 1 when the predicted cluster labels perfectly match the ground-truth labels. Note that the above metrics are computed based only on the local dataset $X_{i,j}$ in the case of local clustering, whereas in DC-Clustering and centralized clustering, they are computed based on the entire dataset X. Therefore, although the metrics obtained from local clustering carry a certain significance, they are treated merely as reference values. The ground-truth cluster labels were defined as the original generative distribution labels for the synthetic data experiments and as the pre-assigned labels in the datasets for the open-data experiments.

\subsection{Experiment I: Proof-of-concept using synthetic data}
\label{sec:experiment_I_proof-of-concept_using_synthetic_data}

\subsubsection{Settings}
\label{sec:experiment_I_settings}
In this study, we conducted a proof-of-concept experiment using synthetic data to intuitively assess the effectiveness of the proposed method. We consider two settings: one in which the data distribution is independent and identically distributed (IID), and the other in which it is non-IID.

\begin{figure}[tbp]
    \centering
    \begin{minipage}[b]{1.0\textwidth}
        \raggedleft
        \includegraphics[width=0.15\textwidth]{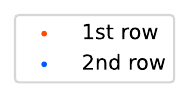}
    \end{minipage}

    \vspace{2pt}

    \begin{subfigure}{0.48\textwidth}
        \centering
        \includegraphics[width=\textwidth]{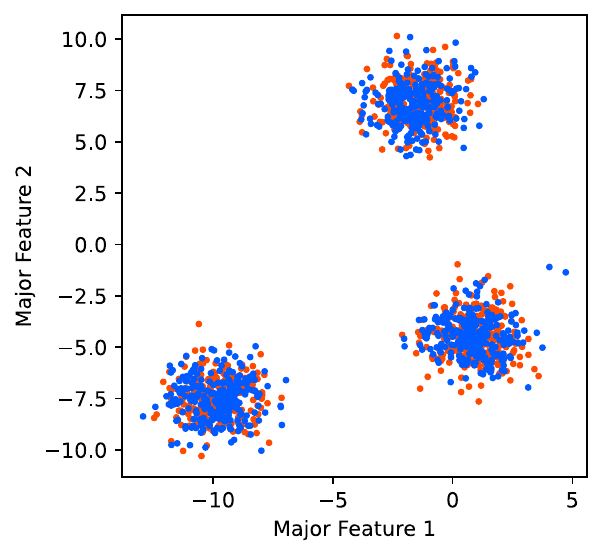}
        \caption{Blobs in IID setting}
    \end{subfigure}
    \hfill
    \begin{subfigure}{0.48\textwidth}
        \centering
        \includegraphics[width=\textwidth]{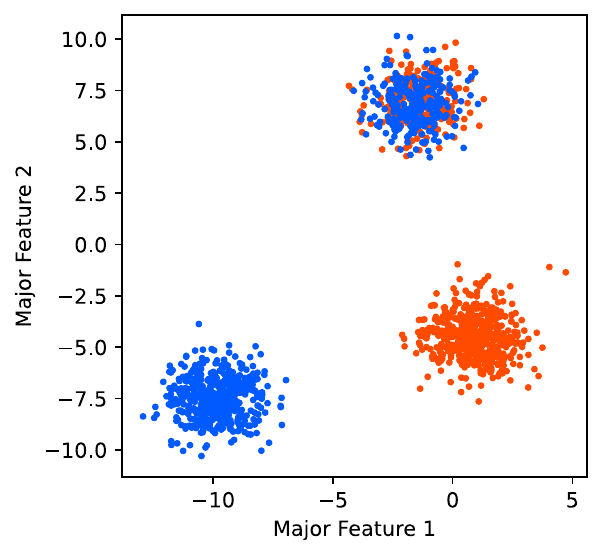}
        \caption{Blobs in non-IID setting}
    \end{subfigure}

    \vspace{10pt} 

    \begin{subfigure}{0.48\textwidth}
        \centering
        \includegraphics[width=\textwidth]{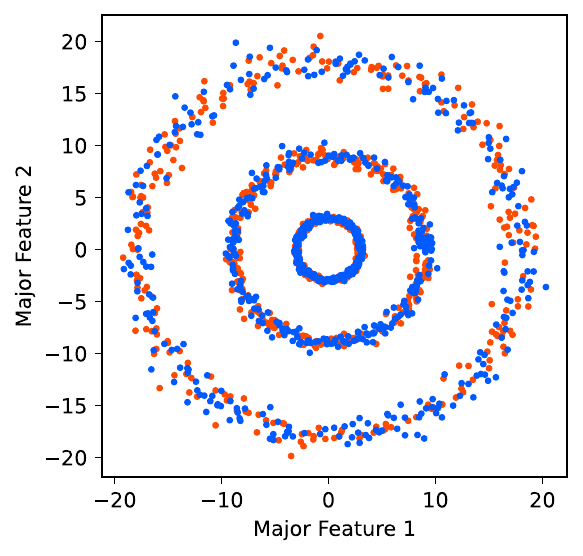}
        \caption{Circles in IID setting}
    \end{subfigure}
    \hfill
    \begin{subfigure}{0.48\textwidth}
        \centering
        \includegraphics[width=\textwidth]{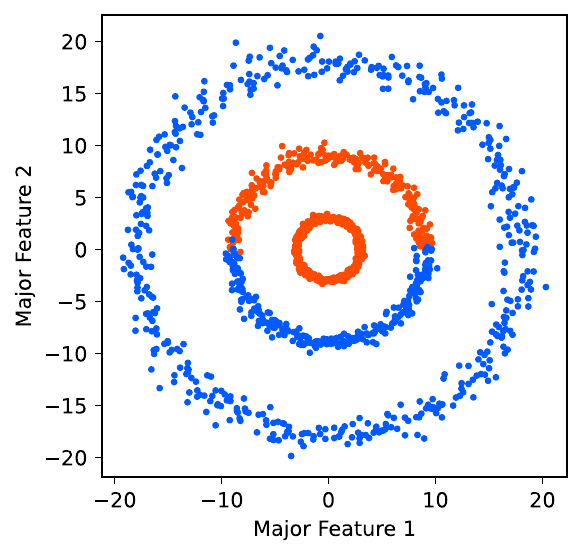}
        \caption{Circles in non-IID setting}
    \end{subfigure}

    \caption{Distribution of generated synthetic data. (a) and (b) represent IID and non-IID setting, respectively, for Blobs, and (c) and (d) represents those for Circles. Orange and blue markers represent datasets with first-row parties and second-row parties, respectively.}
    \label{fig:synthetic_data}
\end{figure}

The synthetic datasets used in this experiment (Fig. \ref{fig:synthetic_data}) consisted of two major features, four minor features, and ground-truth cluster labels. The major features and cluster labels were generated using the $\mathtt{make\_blobs}$ and $\mathtt{make\_circles}$ functions from the scikit-learn library \citep{scikit-learn}. The number of ground-truth clusters was set to three. The minor features were generated from multivariate Gaussian noise with a mean of 0 and a covariance matrix $S \in \mathbb{R}^{4 \times 4}$. The diagonal entries of $S$ were set to 0.1, and the off-diagonal entries were set to 0.01. Synthetic data generated from the Blobs and Circles were used for comparative experiments focusing on k-means clustering and SC. Each cluster contained 500 data points, resulting in 1500 data points.

We assume a combination of vertical and horizontal partitioning for the data distribution. Specifically, the scenario involved four institutions participating in the integrated analysis with $c=2$ and $d=2$. For vertical data partitioning, first-column institutions possess major features 1 and two of the minor features, while second-column institutions possess major feature 2 and the remaining two minor features.

For horizontal data partitioning, we consider two scenarios: one in which the datasets held by the first- and second-row institutions are IID and the other in which they are non-IID. In the IID scenario, the data points are randomly allocated to the first- and second-row institutions (see Fig. \ref{fig:synthetic_data}(a) and \ref{fig:synthetic_data}(c)). In contrast, in the non-IID scenario, the institutions in the first and second columns possess the major features shown in Fig. \ref{fig:synthetic_data}(b) and \ref{fig:synthetic_data}(d), respectively. In this setting, both the first- and second-row institutions have data corresponding to the two clusters, but one of the clusters is not shared with the other. Under such conditions, it is difficult to accurately predict true cluster labels unless the datasets held by all institutions are integrated through collaborative clustering.

In the IID setting of the synthetic data experiment for local clustering, we present only the results for institutions (1,1) and (1,2). This is because the data points were randomly assigned, making the results for institutions (2,1) and (2,2) virtually equivalent to those for (1,1) and (1,2), respectively.

\subsubsection{Results}
\label{sec:experiment_I_results}
We first report the results of comparative experiments focusing on k-means. For the datasets shown in Fig. \ref{fig:synthetic_data}(a) and \ref{fig:synthetic_data}(b), the clustering results obtained using k-means clustering with the proposed method, centralized clustering, and local clustering are summarized in the Blobs section of Table \ref{tab:results_synthetic}. As shown in Table \ref{tab:results_synthetic}, the proposed method achieved a performance identical to that of centralized clustering across all metrics in both IID and non-IID settings. All metrics recorded a perfect score of 1.0, indicating that the predicted cluster labels matched the ground-truth labels exactly. By contrast, local clustering, although provided as a reference, did not achieve such an agreement. These results suggest that the proposed method using k-means is effective for data distribution scenarios such as those illustrated in Figs. \ref{fig:synthetic_data}(a) and \ref{fig:synthetic_data}(b).

\begin{table}[tbp]
    \centering
    \caption{Results of synthetic data experiments.}
    \label{tab:results_synthetic}
    \small
    \begin{tabular}{lllccc}
    \hline
     & & & ARI & NMI & ACC \\ 
    \hline
    \multirow{10}{*}{\rotatebox{90}{Blobs (for k-means)}} & \multirow{4}{*}{IID setting} & Local Clustering (1,1) & 0.786 & 0.777 & 0.920 \\
     & & Local Clustering (1,2) & 0.877 & 0.853 & 0.956 \\
     & & Centralized Clustering & 1.000 & 1.000 & 1.000 \\
     & & \textbf{Proposed method} & \textbf{1.000} & \textbf{1.000} & \textbf{1.000} \\
    \cline{2-6}
     & \multirow{6}{*}{Non-IID setting} & Local Clustering (1,1) & 0.295 & 0.357 & 0.591 \\
     & & Local Clustering (1,2) & 0.571 & 0.761 & 0.673 \\
     & & Local Clustering (2,1) & 0.141 & 0.209 & 0.511 \\
     & & Local Clustering (2,2) & 0.142 & 0.209 & 0.508 \\
     & & Centralized Clustering & 1.000 & 1.000 & 1.000 \\
     & & \textbf{Proposed method} & \textbf{1.000} & \textbf{1.000} & \textbf{1.000} \\
    \hline
    \multirow{10}{*}{\rotatebox{90}{Circle (for SC)}} & \multirow{4}{*}{IID setting} & Local Clustering (1,1) & 0.261 & 0.294 & 0.579 \\
     & & Local Clustering (1,2) & 0.253 & 0.286 & 0.575 \\
     & & Centralized Clustering & 1.000 & 1.000 & 1.000 \\
     & & \textbf{Proposed method} & \textbf{1.000} & \textbf{1.000} & \textbf{1.000} \\
    \cline{2-6}
     & \multirow{6}{*}{Non-IID setting} & Local Clustering (1,1) & 0.261 & 0.294 & 0.579 \\
     & & Local Clustering (1,2) & 0.253 & 0.286 & 0.575 \\
     & & Local Clustering (2,1) & -0.001 & 0.001 & 0.351 \\
     & & Local Clustering (2,2) & -0.002 & 0.001 & 0.348 \\
     & & Centralized Clustering & 1.000 & 1.000 & 1.000 \\
     & & \textbf{Proposed method} & \textbf{1.000} & \textbf{1.000} & \textbf{1.000} \\
    \hline
    \end{tabular}
\end{table}

Next, we report the results of the comparative experiments focusing on SC. For the datasets shown in Figs. \ref{fig:synthetic_data}(c) and \ref{fig:synthetic_data}(d), the clustering results obtained using SC with the proposed method, centralized clustering, and local clustering are presented in the circular section of Table 2. Similar to the k-means results, the proposed method using SC demonstrated comparable performance. These findings indicate that the proposed method using SC is effective even for complex data structures such as those illustrated in Fig. \ref{fig:synthetic_data}(c) and \ref{fig:synthetic_data}(d).

The results from the synthetic data experiments demonstrate that the proposed method can achieve a performance comparable to centralized clustering, even when the information necessary to predict the ground-truth cluster labels within the local institutions’ datasets is insufficient. However, it should be noted that these results are based on a single trial conducted for proof-of-concept purposes and merely serve as an illustrative example of the effectiveness of the proposed method. In the next section, we examine the practicality of the proposed method through a comprehensive evaluation of open datasets.

\subsection{Experiment II: Evaluation performance using open data}
\label{sec:experiment_II_evaluation_performance_using_open_data}

\subsubsection{Settings}
\label{sec:experiment_II_settings}
In this experiment, we evaluate the practical applicability of DC-Clustering using open datasets. Similar to the synthetic data experiment, the baselines were centralized and locally clustered. As mentioned in the introduction, no existing methods that can handle scenarios that combine both vertical and horizontal partitioning have been proposed. Therefore, comparisons with methods that support only horizontal partitioning are expected to provide valuable insights. For comparison purposes, we selected two pioneering studies that extended k-means and spectral clustering to a federated clustering framework: \cite{kumar2020federated} and \cite{hernandez2021federated}.

The experiment used six open datasets that have been employed in previous studies, such as those of \cite{barcena2024federated} and \cite{qiao2023federated}. These datasets include Iris \citep{fisher1936use}, Rice \citep{cinar2019rice}, Pendigits \citep{alpaydin1998pen}, Heart-statlog \citep{aha1988statlog}, and Bank \citep{lohweg2012banknote} from the UCI Machine Learning Repository \citep{kellyUCI}, and Phoneme from the KEEL-dataset repository \citep{alcala2011keel}. The number of ground-truth clusters, sample sizes, and features for each dataset are summarized in Table \ref{tb:datasets}.

\begin{table}[tbp]
    \centering
    \caption{Details of open datasets.}
    \label{tab:dataset_details}
    \begin{tabular}{lrrrrr}
    \hline
    Dataset & $n$ & $m$ & GT clust. num. & GT min. clust. size. & GT max. clust. size. \\ 
    \hline
    Iris & 150 & 4 & 3 & 50 & 50 \\
    Rice & 3810 & 7 & 2 & 1630 & 2180 \\
    Pendigits & 10992 & 16 & 10 & 1054 & 1144 \\
    Heart-statlog & 270 & 13 & 2 & 120 & 150 \\
    Bank & 1372 & 4 & 2 & 1471 & 3529 \\
    Phoneme & 5404 & 5 & 2 & 608 & 762 \\
    \hline
    \end{tabular}
    \label{tb:datasets}
\end{table}

For the data-partitioning setting, we set $c=10$ and $d=2$. The sample size $n_{i,j}$ for each institution $(i,j)$ was set by dividing the total sample size equally across the $c$ rows such that each institution held the same number of data points. The number of features $m_{i,j}$ is similarly divided evenly across the $d$ columns. To simplify the experimental design, the data points assigned to each row-wise institution were randomly selected for each trial, and the features held by each column-wise institution were randomly allocated per trial. For the methods of \cite{kumar2020federated} and \cite{hernandez2021federated}, clustering was performed using only institutions located at positions (1,1),(2,1), $\cdots$, (10,1).

Additionally, for each dataset, comparative experiments were conducted using the k-means and SC methods. This was performed to verify the effectiveness of the proposed method with respect to each clustering algorithm, reflecting its flexibility in selecting and applying clustering methods according to the analytical objectives.

Evaluation metrics were calculated for each trial for the performance evaluation in the open-data experiment, and the average and standard deviation over 100 trials were reported. This number of trials is greater than those in most previous studies on federated clustering and is expected to provide reliable results. In each trial, factors such as the assignment of data to institutions and random seeds for k-means varied. For local clustering, we present only the results for institution (1,1). This is because, owing to the random assignment of data points and features, the results of institution (1,1) are essentially equivalent to those of other institutions. The number of federation steps in \cite{kumar2020federated} was set to 5 in the original study; however, in this experiment, it was increased to 10 to obtain more stable results.

\subsubsection{Results}
\label{sec:experiment_II_results}
First, we present the results of a comparative experiment focusing on k-means. Table \ref{tb:results_expII_kmeans} shows the results of applying the proposed method, centralized clustering, local clustering, and the method of \cite{kumar2020federated} to each dataset using k-means. 
The proposed method slightly underperformed central clustering on all metrics for Pendigits and Bank, as well as on the NMI for Phoneme, but the percentage difference in absolute values is at most 4.6\%. Furthermore, Table \ref{tb:results_exp_summary}, which summarizes the average differences in metrics between central clustering and each method overall, shows values of 1.2\%, 31.6\%, and 48.2\% for the proposed method, \cite{kumar2020federated}, and local clustering, respectively. This indicates that the proposed method demonstrates relatively comparable performance to central clustering in k-means.

\begin{table}[tbp]
    \centering
    \caption{Results of Experiment II for k-means.}
    \resizebox{\textwidth}{!}{
    \setlength{\tabcolsep}{3pt}
    \begin{tabular}{ll cc r cc r cc r}
    \hline
     & & \multicolumn{2}{c}{ARI} & \multicolumn{1}{c}{\% $\Delta$ ARI} & \multicolumn{2}{c}{NMI} & \multicolumn{1}{c}{\% $\Delta$ NMI} & \multicolumn{2}{c}{ACC} & \multicolumn{1}{c}{\% $\Delta$ ACC} \\
     & & \multicolumn{2}{c}{} & \multicolumn{1}{c}{vs. CC} & \multicolumn{2}{c}{} & \multicolumn{1}{c}{vs. CC} & \multicolumn{2}{c}{} & \multicolumn{1}{c}{vs. CC} \\
    \hline
    \multirow{4}{*}{\rotatebox{90}{Iris}} & Local Clustering (1,1) & 0.683 & (0.245) & 6.4\% & 0.773 & (0.169) & 2.0\% & 0.854 & (0.124) & 4.4\% \\
     & Kumar et al.(2020) & 0.735 & (0.105) & 0.7\% & 0.737 & (0.085) & 2.8\% & 0.893 & (0.051) & 0.1\% \\
     & \textbf{The proposed method} & \textbf{0.752} & \textbf{(0.015)} & 2.9\% & \textbf{0.775} & \textbf{(0.013)} & 2.3\% & \textbf{0.904} & \textbf{(0.007)} & 1.2\% \\
     & Central Clustering & 0.730 & (0.001) & & 0.758 & (0.002) & & 0.893 & (0.001) & \\
    \hline
    \multirow{4}{*}{\rotatebox{90}{Rice}} & Local Clustering (1,1) & 0.599 & (0.111) & 3.8\% & 0.497 & (0.099) & 6.1\% & 0.885 & (0.045) & 0.6\% \\
     & Kumar et al.(2020) & 0.595 & (0.105) & 3.1\% & 0.489 & (0.092) & 4.2\% & 0.883 & (0.044) & 0.4\% \\
     & \textbf{The proposed method} & \textbf{0.577} & \textbf{(0.000)} & 0.0\% & \textbf{0.469} & \textbf{(0.000)} & 0.0\% & \textbf{0.880} & \textbf{(0.000)} & 0.0\% \\
     & Central Clustering & 0.577 & (0.000) & & 0.469 & (0.000) & & 0.880 & (0.000) & \\
    \hline
    \multirow{4}{*}{\rotatebox{90}{Pendigits}} & Local Clustering (1,1) & 0.455 & (0.068) & 19.0\% & 0.605 & (0.051) & 11.4\% & 0.625 & (0.068) & 12.1\% \\
     & Kumar et al.(2020) & 0.443 & (0.058) & 21.1\% & 0.586 & (0.046) & 14.3\% & 0.608 & (0.058) & 14.4\% \\
     & \textbf{The proposed method} & \textbf{0.548} & \textbf{(0.034)} & 2.4\% & \textbf{0.678} & \textbf{(0.012)} & 0.8\% & \textbf{0.696} & \textbf{(0.046)} & 2.1\% \\
     & Central Clustering & 0.561 & (0.033) & & 0.683 & (0.010) & & 0.711 & (0.048) & \\
    \hline
    \multirow{4}{*}{\rotatebox{90}{Heart-statlog}} & Local Clustering (1,1) & 0.045 & (0.098) & 49.0\% & 0.064 & (0.074) & 223.8\% & 0.623 & (0.077) & 5.1\% \\
     & Kumar et al.(2020) & 0.057 & (0.066) & 88.5\% & 0.041 & (0.051) & 108.1\% & 0.610 & (0.056) & 3.0\% \\
     & \textbf{The proposed method} & \textbf{0.030} & \textbf{(0.000)} & 0.1\% & \textbf{0.020} & \textbf{(0.000)} & 0.1\% & \textbf{0.593} & \textbf{(0.000)} & 0.0\% \\
     & Central Clustering & 0.030 & (0.000) & & 0.020 & (0.000) & & 0.593 & (0.000) & \\
    \hline
    \multirow{4}{*}{\rotatebox{90}{Bank}} & Local Clustering (1,1) & 0.129 & (0.133) & 169.8\% & 0.099 & (0.108) & 231.2\% & 0.666 & (0.082) & 8.9\% \\
     & Kumar et al.(2020) & 0.096 & (0.125) & 99.9\% & 0.071 & (0.102) & 138.3\% & 0.639 & (0.071) & 4.5\% \\
     & \textbf{The proposed method} & \textbf{0.047} & \textbf{(0.009)} & 2.1\% & \textbf{0.029} & \textbf{(0.006)} & 2.3\% & \textbf{0.610} & \textbf{(0.010)} & 0.2\% \\
     & Central Clustering & 0.048 & (0.000) & & 0.030 & (0.000) & & 0.612 & (0.000) & \\
    \hline
    \multirow{4}{*}{\rotatebox{90}{Phoneme}} & Local Clustering (1,1) & 0.035 & (0.096) & 67.5\% & 0.116 & (0.043) & 37.8\% & 0.611 & (0.068) & 8.5\% \\
     & Kumar et al.(2020) & 0.080 & (0.120) & 25.8\% & 0.121 & (0.043) & 35.3\% & 0.639 & (0.084) & 4.2\% \\
     & \textbf{The proposed method} & \textbf{0.108} & \textbf{(0.011)} & 0.2\% & \textbf{0.178} & \textbf{(0.014)} & 4.6\% & \textbf{0.667} & \textbf{(0.007)} & 0.1\% \\
     & Central Clustering & 0.107 & (0.002) & & 0.186 & (0.001) & & 0.667 & (0.001) & \\
    \hline
    \multicolumn{11}{p{\textwidth}}{\footnotesize The metric values are averages and standard deviations in parentheses for 100 trials. ``\% $\Delta$ [Metric] vs. CC'' denotes the absolute percentage difference in a given average metric (ARI, NMI, ACC) compared to central clustering (CC).}
    \end{tabular}
    }
    \label{tb:results_expII_kmeans}
\end{table}

\begin{table}[tbp]
    \centering
    \caption{Average percentage differences compared to central clustering (CC).}
    \begin{tabular}{l rr}
    \hline
     & \multicolumn{2}{c}{Average of ``\% $\Delta$ [Metric] vs. CC''} \\
    \cline{2-3} 
     & \multicolumn{1}{c}{k-means} & \multicolumn{1}{c}{SC} \\ 
    \hline
    Local Clustering (1,1) & 48.2\% & 139.0\% \\
    Kumar et al.(2020) & 31.6\% & \\
    Hernández-Pereira et al. (2021) & & 59.0\% \\
    \textbf{The proposed method} & \textbf{1.2\%} & \textbf{15.0\%} \\
    \hline
    \multicolumn{3}{p{0.5\linewidth}}{ 
        {\footnotesize Averages of absolute percentage differences in benchmark datasets and metrics (ARI, NMI, ACC) compared to central clustering (CC).}
    }
    \end{tabular}
    \label{tb:results_exp_summary}
\end{table}

Although presented only as reference values, local clustering produced better results than centralized clustering in some cases, specifically for all metrics on the Rice, Heart-statlog, and Bank datasets, and for NMI on the Iris dataset. However, the standard deviations of these results were notably large, indicating instability. Considering this, from the perspective of achieving stable performance comparable to centralized clustering, it is more effective for local institutions to use the proposed method than to apply k-means solely to their local data.

The results of applying SC to the proposed method, centralized clustering, local clustering, and the method proposed by \cite{hernandez2021federated} for each dataset are listed in Table \ref{tb:results_expII_sc}. The proposed method slightly underperformed in terms of ARI and NMI for the Rice dataset and across all metrics for the Phoneme dataset; however, the maximum relative difference in absolute values was at most 1.9\%. Furthermore, as shown in Table \ref{tb:results_exp_summary}, the average differences between the metrics of centralized clustering were 15.0\% for the proposed method, 59.0\% for \cite{hernandez2021federated}’s method, and 139.0\% for local clustering. These results indicate that, in the context of SC, the proposed method demonstrates performance comparable to that of centralized clustering.

\begin{table}[tbp]
    \centering
    \caption{Results of Experiment II for SC. Same format and interpretation as Table 4.}
    \resizebox{\textwidth}{!}{
    \setlength{\tabcolsep}{3pt}
    \begin{tabular}{ll cc r cc r cc r}
    \hline
     & & \multicolumn{2}{c}{ARI} & \multicolumn{1}{c}{\% $\Delta$ ARI} & \multicolumn{2}{c}{NMI} & \multicolumn{1}{c}{\% $\Delta$ NMI} & \multicolumn{2}{c}{ACC} & \multicolumn{1}{c}{\% $\Delta$ ACC} \\
     & & \multicolumn{2}{c}{} & \multicolumn{1}{c}{vs. CC} & \multicolumn{2}{c}{} & \multicolumn{1}{c}{vs. CC} & \multicolumn{2}{c}{} & \multicolumn{1}{c}{vs. CC} \\
    \hline
    \multirow{4}{*}{\rotatebox{90}{Iris}} & Local Clustering (1,1) & 0.464 & (0.185) & 38.9\% & 0.610 & (0.143) & 24.3\% & 0.755 & (0.106) & 16.7\% \\
     & Hernández-Pereira et al. (2021) & 0.008 & (0.076) & 99.0\% & 0.021 & (0.080) & 97.4\% & 0.392 & (0.056) & 56.8\% \\
     & \textbf{The proposed method} & \textbf{0.787} & \textbf{(0.052)} & 3.6\% & \textbf{0.806} & \textbf{(0.032)} & 0.0\% & \textbf{0.918} & \textbf{(0.024)} & 1.3\% \\
     & Central Clustering & 0.759 & (0.000) & & 0.806 & (0.000) & & 0.907 & (0.000) & \\
    \hline
    \multirow{4}{*}{\rotatebox{90}{Rice}} & Local Clustering (1,1) & 0.517 & (0.140) & 0.3\% & 0.443 & (0.114) & 4.7\% & 0.856 & (0.056) & 0.5\% \\
     & Hernández-Pereira et al. (2021) & 0.005 & (0.053) & 99.0\% & 0.004 & (0.043) & 98.9\% & 0.511 & (0.036) & 40.6\% \\
     & \textbf{The proposed method} & \textbf{0.518} & \textbf{(0.017)} & 0.2\% & \textbf{0.422} & \textbf{(0.011)} & 0.2\% & \textbf{0.860} & \textbf{(0.006)} & 0.1\% \\
     & Central Clustering & 0.519 & (0.014) & & 0.423 & (0.010) & & 0.860 & (0.005) & \\
    \hline
    \multirow{4}{*}{\rotatebox{90}{Pendigits}} & Local Clustering (1,1) & 0.526 & (0.064) & 6.8\% & 0.694 & (0.043) & 11.6\% & 0.681 & (0.057) & 6.1\% \\
     & Hernández-Pereira et al. (2021) & 0.006 & (0.056) & 99.0\% & 0.010 & (0.078) & 98.8\% & 0.119 & (0.061) & 83.6\% \\
     & \textbf{The proposed method} & \textbf{0.592} & \textbf{(0.052)} & 5.0\% & \textbf{0.796} & \textbf{(0.017)} & 1.5\% & \textbf{0.743} & \textbf{(0.030)} & 2.5\% \\
     & Central Clustering & 0.564 & (0.001) & & 0.785 & (0.001) & & 0.725 & (0.001) & \\
    \hline
    \multirow{4}{*}{\rotatebox{90}{Heart-statlog}} & Local Clustering (1,1) & 0.039 & (0.094) & 20.6\% & 0.059 & (0.071) & 72.1\% & 0.611 & (0.079) & 0.5\% \\
     & Hernández-Pereira et al. (2021) & 0.001 & (0.007) & 98.5\% & 0.003 & (0.005) & 90.9\% & 0.527 & (0.021) & 14.3\% \\
     & \textbf{The proposed method} & \textbf{0.049} & \textbf{(0.000)} & 0.1\% & \textbf{0.034} & \textbf{(0.000)} & 0.1\% & \textbf{0.615} & \textbf{(0.001)} & 0.0\% \\
     & Central Clustering & 0.049 & (0.000) & & 0.034 & (0.000) & & 0.615 & (0.000) & \\
    \hline
    \multirow{4}{*}{\rotatebox{90}{Bank}} & Local Clustering (1,1) & 0.163 & (0.150) & 1919.5\% & 0.163 & (0.145) & 223.1\% & 0.686 & (0.090) & 20.2\% \\
     & Hernández-Pereira et al. (2021) & 0.004 & (0.000) & 52.0\% & 0.035 & (0.000) & 31.2\% & 0.564 & (0.000) & 1.2\% \\
     & \textbf{The proposed method} & \textbf{0.024} & \textbf{(0.030)} & 202.0\% & \textbf{0.074} & \textbf{(0.017)} & 46.8\% & \textbf{0.589} & \textbf{(0.029)} & 3.3\% \\
     & Central Clustering & 0.008 & (0.004) & & 0.050 & (0.014) & & 0.570 & (0.006) & \\
    \hline
    \multirow{4}{*}{\rotatebox{90}{Phoneme}} & Local Clustering (1,1) & 0.028 & (0.096) & 84.9\% & 0.090 & (0.051) & 37.8\% & 0.617 & (0.077) & 14.0\% \\
     & Hernández-Pereira et al. (2021) & 0.187 & (0.000) & 0.1\% & 0.144 & (0.000) & 0.1\% & 0.717 & (0.000) & 0.0\% \\
     & \textbf{The proposed method} & \textbf{0.183} & \textbf{(0.009)} & 1.9\% & \textbf{0.143} & \textbf{(0.004)} & 1.0\% & \textbf{0.715} & \textbf{(0.006)} & 0.3\% \\
     & Central Clustering & 0.187 & (0.000) & & 0.144 & (0.000) & & 0.717 & (0.000) & \\
    \hline
    \end{tabular}
    }
    \label{tb:results_expII_sc}
\end{table}

Moreover, for the Bank dataset across all metrics and for NMI in the Rice and Heart-statlog datasets, local clustering outperformed centralized clustering. However, similar to the case of k-means, the standard deviations of these results were notably large. These findings suggest that even in the case of SC, utilizing the proposed method is effective for individual institutions.

Based on the results of the open-data experiments, it was confirmed that the proposed method consistently achieved clustering accuracy comparable to that of centralized clustering for both k-means and SC. In particular, by avoiding the instability observed when individual institutions perform clustering based solely on their own data, and by accurately capturing the overall data structure, the proposed method can be regarded as a practical and reliable clustering approach for distributed data environments.

\section{Discussion}
\label{sec:discussion}
This section examines the validity of the experimental results from four perspectives, based on \cite{feldt2010validity}: credibility, internal validity, construct validity, and external validity. In addition, future research directions are discussed.

Credibility evaluates whether the results obtained are accurate and reliable. In this study, to reduce human error and enhance the reproducibility and reliability of the results, different authors implemented and reviewed the experimental code. This approach minimized the likelihood of errors in the numerical experimentation process.

Internal validity concerns whether the observed effects can be attributed to the proposed method. In the synthetic data experiments, a simple setting with minimal partitioning ($c=2,d=2$) was adopted to eliminate the influence of extraneous factors and to verify the fundamental performance of the method. In open-data experiments, multiple trials were conducted using different random seeds to assess the influence of randomness and confirm the consistency of the results.

Construct validity assesses whether the theoretical foundation of a study aligns with the observed outcomes. The proposed method achieved performance comparable to that of centralized clustering by integrating data points and features distributed across institutions. This result is a natural consequence of the fundamental concept behind the proposed method, in which distributed information can be effectively aggregated through the sharing of intermediate representations.

External validity refers to the extent to which the results of a study can be generalized to different situations or datasets. The proposed method was evaluated using two types of synthetic data and six types of open datasets. However, its effectiveness on other datasets remains unverified, which is a limitation of the present study. For example, in the case of high-dimensional data \citep{yata2010effective} that exhibit characteristics not considered in this study, the proposed method may not necessarily perform effectively, and further refinement of the method may be required.

Finally, future research directions include automatic determination of the number of clusters\citep{mur2016determination}, extensions to hierarchical or soft clustering, and clustering methods targeting graph-structured data \citep{abdalameer2022new}. In addition, the introduction of visualization techniques to present the resulting cluster structures intuitively is a valuable direction for exploration.

\section{Conclusion}
\label{sec:conclusion}
In this paper, we propose DC-Clustering as a method for clustering distributed data while preserving privacy. The proposed method preserved the privacy of each institution by sharing intermediate representations instead of raw data. A key feature of this method is its ability to perform integrated clustering under complex data distribution settings, where horizontal and vertical partitioning are combined, which existing methods cannot handle. Furthermore, it supports the flexible use of both k-means and SC depending on the user’s analytical objectives and completes the communication process through a single round of interaction. Numerical experiments using both synthetic and open datasets demonstrate that the performance of the proposed method is comparable to that of centralized clustering. Based on these results, DC-Clustering is expected to serve as an effective clustering approach in distributed data environments and to contribute to future research on federated data analysis and privacy-preserving machine learning. Additionally, the proposed method offers potential as a foundational technology for knowledge discovery in distributed settings because it enables the secure integration of diverse data held by each institution and reveals the latent structures among them.


\section*{Acknowledgement}
This work was supported in part by the Japan Society for the Promotion of Science (JSPS) and Japan Grants-in-Aid for Scientific Research (No. JP23K22166). The authors would like to thank Editage (www.editage.jp) for the English language editing.

\section*{Declaration of generative AI and AI-assisted technologies in the writing process}
During the preparation of this work the authors used ChatGPT (OpenAI) to assist with phrasing, grammar correction, and refinement of English expressions in the manuscript. After using this tool, the authors reviewed and edited the content as needed and take full responsibility for the content of the publication.

\appendix

\section{A justification of affine transform}
\label{app:affine}
When $f_i$ is an affine transformation (e.g., PCA) rather than a purely linear transformation, it is intuitively expected that assuming $g_i$ as an affine transformation leads to a more accurate approximation in (\ref{eq:g_approx}).
This appendix provides theoretical support for this intuition. However, the following discussion does not necessarily claim that assuming affine transformations yields better DC-Clustering results than assuming linear transformations. 
In the following, the affine transformation $g_i$ is formally expressed as $g_i = (G_i^\prime, G_i^*)$. 
In addition, we limit our discussion to the case in which $\tilde{m}_i = \tilde{m}$ for all $i=1,\cdots,c$.

First, we consider the case in which $c=2$. Functions $g_1$ and $g_2$ are expected to approximately satisfy (\ref{eq:g_equal}).
\begin{equation}
    g_1(f_1(X^{\text{anc}})) = g_2(f_2(X^{\text{anc}})). \label{eq:g_equal}
\end{equation}
If $g_i$ is assumed to be a linear transformation, the existence of a pair $(g_1,g_2)$ that satisfies (\ref{eq:g_equal}) is not necessarily guaranteed. However, if $g_i$ is assumed to be an affine transformation, it can be shown that, under certain conditions, there exists a pair $(g_1,g_2)$ satisfying (\ref{eq:g_equal}), as demonstrated in Theorem \ref{thm:existence}. Note that the intermediate representation function $f_i$ is represented as a pair $(F_i, \mu_i )$, where $F_i$ is a linear transformation and $\mu_i$ is an average vector.
\begin{thm}
    Suppose that $f_i=(F_i,\mu_i )$ for $i=1,2$ satisfies the following conditions:
    \begin{equation}
        \mathcal{R}(F_1) = \mathcal{R}(F_2), \quad \text{Rank}(X^{\text{anc}} F_1) = \text{Rank}(X^{\text{anc}} F_2) = \widetilde{m},
    \end{equation}
    where $\mathcal{R}$ denotes the range. Then, there exists a pair of affine transformations $(g_1,g_2 )$ that satisfies (\ref{eq:g_equal}).
    \label{thm:existence}
\end{thm}
\begin{proof}
    First, we expand the collaborative representation as follows:
    \begin{align}
        g_1(f_1(X^{\text{anc}})) &= X^{\text{anc}} F_1 G'_1 + \mathbf{1}(G^*_1 - \mu^\top_1 F_1 G'_1), \\
        g_2(f_2(X^{\text{anc}})) &= X^{\text{anc}} F_2 G'_2 + \mathbf{1}(G^*_2 - \mu^\top_2 F_2 G'_2).
    \end{align}
    Given the assumption, there exists a pair $(G'_1, G'_2)$ such that $X^{\text{anc}} F_1 G'_1 = X^{\text{anc}} F_2 G'_2$. Furthermore, by defining $\boldsymbol{G}^*_1 = \mu^\top_1 F_1 G'_1$ and $\boldsymbol{G}^*_2 = \mu^\top_2 F_2 G'_2$ using such $(G'_1, G'_2)$, the corresponding pair $(g_1 = (G'_1, \boldsymbol{G}^*_1), g_2 = (G'_2, \boldsymbol{G}^*_2))$ satisfies (\ref{eq:g_equal}).
\end{proof}
Theorem \ref{thm:existence} can be readily extended to the case where $c>2$ to satisfy (\ref{eq:g_equal_all}),
\begin{equation}
    g_1(f_1(X^{\text{anc}})) = g_2(f_2(X^{\text{anc}})) = \dots = g_c(f_c(X^{\text{anc}})), \label{eq:g_equal_all}
\end{equation}
as shown in Corollary \ref{cor:existence}.
\begin{cor}
    Assume that $f_i = (F_i, \mu_i)$ for $i = 1, 2, \dots, c$ satisfy the following condition:
    \begin{equation}
        \mathcal{R}(F_1) = \mathcal{R}(F_2) = \dots = \mathcal{R}(F_c), \quad \text{Rank}(X^{\text{anc}} F_i) = \widetilde{m}.
    \end{equation}
    Then, there exist affine transformations $(g_1, g_2, \dots, g_c)$ that satisfy (\ref{eq:g_equal_all}).
    \label{cor:existence}
\end{cor}
\begin{proof}
Given the assumption, there exists a set of matrices $(G'_1, G'_2, \dots, G'_c)$ such that $F_1 G'_1 = F_2 G'_2 = \dots = F_c G'_c$. Based on this, consider $\mathbf{G}^*_i = \mu^\top_i F_i G'_i$ for each $i$. The set of affine transformations $g_i = (G'_i, \mathbf{G}^*_i)$ constructed in this manner clearly satisfies (\ref{eq:g_equal_all}), based on the proof of Theorem \ref{thm:existence}.
\end{proof}

If the conditions of Corollary \ref{cor:existence} are satisfied, and a transformation $g_i$ that strictly satisfies (\ref{eq:g_approx}) can be obtained, more accurate clustering results can be expected. In fact, our preliminary numerical experiments confirmed that assuming an affine transformation for $g_i$ led to improved clustering performance compared to assuming a linear transformation. However, it should be noted that the method described in the Methods section does not always guarantee the construction of such a transformation $g_i$, and that these discussions are based on anchor data $X^\text{anc}$. Therefore, it does not necessarily follow that high-performance clustering results will be obtained for the target dataset $X$. Further research is needed to clarify the conditions and contexts under which the affine assumption for $g_i$ is more effective than the linear assumption.

\section{Glossary}
\label{app:glossary}
\small

\begin{xltabular}{\textwidth}{|l|X|}
    
    \caption{Term and Definition} \label{tab:terms} \\
    
    \hline
    \textbf{Term} & \textbf{Definition} \\ \hline
    \endfirsthead

    \hline
    \textbf{Term} & \textbf{Definition} (Continued) \\ \hline
    \endhead

    \hline
    \multicolumn{2}{|r|}{{Continued on next page}} \\ \hline
    \endfoot

    \hline
    \endlastfoot

    ACC (Clustering accuracy) & The proportion of correctly clustered data points based on optimal label permutation between predicted and ground-truth clusters. \\ \hline
    Affine transformation & A linear transformation followed by a translation, used in this study for constructing intermediate and collaborative representations. \\ \hline
    Anchor dataset & A shared dataset used to construct transformation functions for collaborative representations across institutions. \\ \hline
    ARI (Adjusted Rand index) & A clustering evaluation metric measuring the similarity between predicted and true cluster labels, adjusted for chance. \\ \hline
    Collaborative representation & A unified representation constructed by the analyst from intermediate representations, used for clustering without exposing original data. \\ \hline
    DC (Data collaboration) analysis & A non-model-sharing approach to federated learning in which each institution shares dimensionally reduced intermediate representations instead of model parameters. \\ \hline
    DC-Clustering & The proposed method in this study; a privacy-preserving federated clustering framework based on DC analysis, supporting both k-means and spectral clustering. \\ \hline
    Federated clustering & A branch of federated learning focused on performing clustering tasks over distributed datasets while preserving privacy. \\ \hline
    FL (Federated learning) & A machine learning framework that enables training across decentralized data sources without exchanging raw data. \\ \hline
    Horizontal partitioning & A data distribution scenario where each institution holds different data points with the same set of features. \\ \hline
    IID (Independent and identically distributed) & A condition where data points are independently sampled from the same distribution. \\ \hline
    Intermediate representation & A dimensionally reduced form of local data generated by each institution, which is shared with a central analyst. \\ \hline
    $k$-means & A widely used clustering algorithm that partitions data into $k$ clusters by minimizing the within-cluster variance. In this paper, ``k-means'' refers to the k-means++ algorithm, which improves centroid initialization to enhance clustering performance and stability. \\ \hline
    NMI (Normalized mutual information) & A metric that evaluates clustering results by comparing the mutual information between predicted and true labels, normalized by their entropy. \\ \hline
    PCA (Principal component analysis) & A dimensionality reduction technique that transforms data to a new coordinate system where the greatest variance lies along the first axes (principal components). PCA is often used to reduce noise and simplify data structures before analysis. \\ \hline
    SC (Spectral clustering) & A clustering method that uses graph Laplacians and eigenvectors to embed data into a lower-dimensional space before applying standard clustering algorithms. \\ \hline
    SVD (Singular value decomposition) & A matrix factorization method that decomposes a matrix into the product of three matrices: $U \Sigma V^T$. It forms the mathematical basis for PCA and is commonly used in signal processing and dimensionality reduction. \\ \hline
    Vertical Partitioning & A data distribution scenario where each institution holds different features for the same set of data points. \\

\end{xltabular}

\bibliography{ref}

\end{document}